\documentclass[10pt,twocolumn,letterpaper]{article}

\usepackage{cvpr}              %

\newcommand{\ours}{Sensor2Sensor\xspace}

\definecolor{cvprblue}{rgb}{0.21,0.49,0.74}
\definecolor{waymogreen}{HTML}{00E89D} %
\definecolor{waymoblue}{HTML}{0077FF} %
\definecolor{waymopurple}{HTML}{9150C8} %
\definecolor{waymoamber}{HTML}{FFCD55} %
\usepackage[pagebackref,breaklinks,colorlinks,allcolors=cvprblue]{hyperref}
\usepackage{caption}
\usepackage{graphicx}
\usepackage{cuted}     %
\usepackage{caption}   %
\usepackage{arydshln}
\usepackage{xspace}
\usepackage{amsmath}
\usepackage[accsupp]{axessibility}
\def\paperID{*****} %
\def\confName{CVPR}
\def\confYear{2026}

\title{\ours: Cross-Embodiment Sensor Conversion for Autonomous Driving}

\author{First Author\\
Institution1\\
Institution1 address\\
{\tt\small firstauthor@i1.org}
\and
Second Author\\
Institution2\\
First line of institution2 address\\
{\tt\small secondauthor@i2.org}
}

\author{Jiahao Wang$^{1,2 \dagger}$, Bo Sun$^1$, Yijing Bai$^1$, Vincent Casser$^1$, Songyou Peng$^3$, Zehao Zhu$^1$, Meng-Li Shih$^{1,4 \dagger}$,\\
Xander Masotto$^1$, Shih-Yang Su$^1$, Kanaad Parvate$^1$, Tiancheng Ge$^1$, Linn Bieske$^1$,\\
Dragomir Anguelov$^1$, Mingxing Tan$^1$, Chiyu Max Jiang$^1$\\
{\small $^1$Waymo, $^2$Johns Hopkins University, $^3$Google DeepMind, $^4$University of Washington}
}
\begin{document}
\maketitle

\begingroup
\renewcommand{\thefootnote}{$\dagger$}
\footnotetext{\footnotesize Work done during an internship at Waymo.}
\endgroup

\addtolength{\abovecaptionskip}{-0.6em} %
\addtolength{\belowcaptionskip}{-0.2em} %
\addtolength{\textfloatsep}{-0.4em} %
\addtolength{\intextsep}{-0.1em} %
\addtolength{\floatsep}{-0.6em} %

\begin{strip}
\vspace{-5.2em}
    \centering
    
    \captionsetup{type=figure}
    \includegraphics[width=0.9\textwidth]{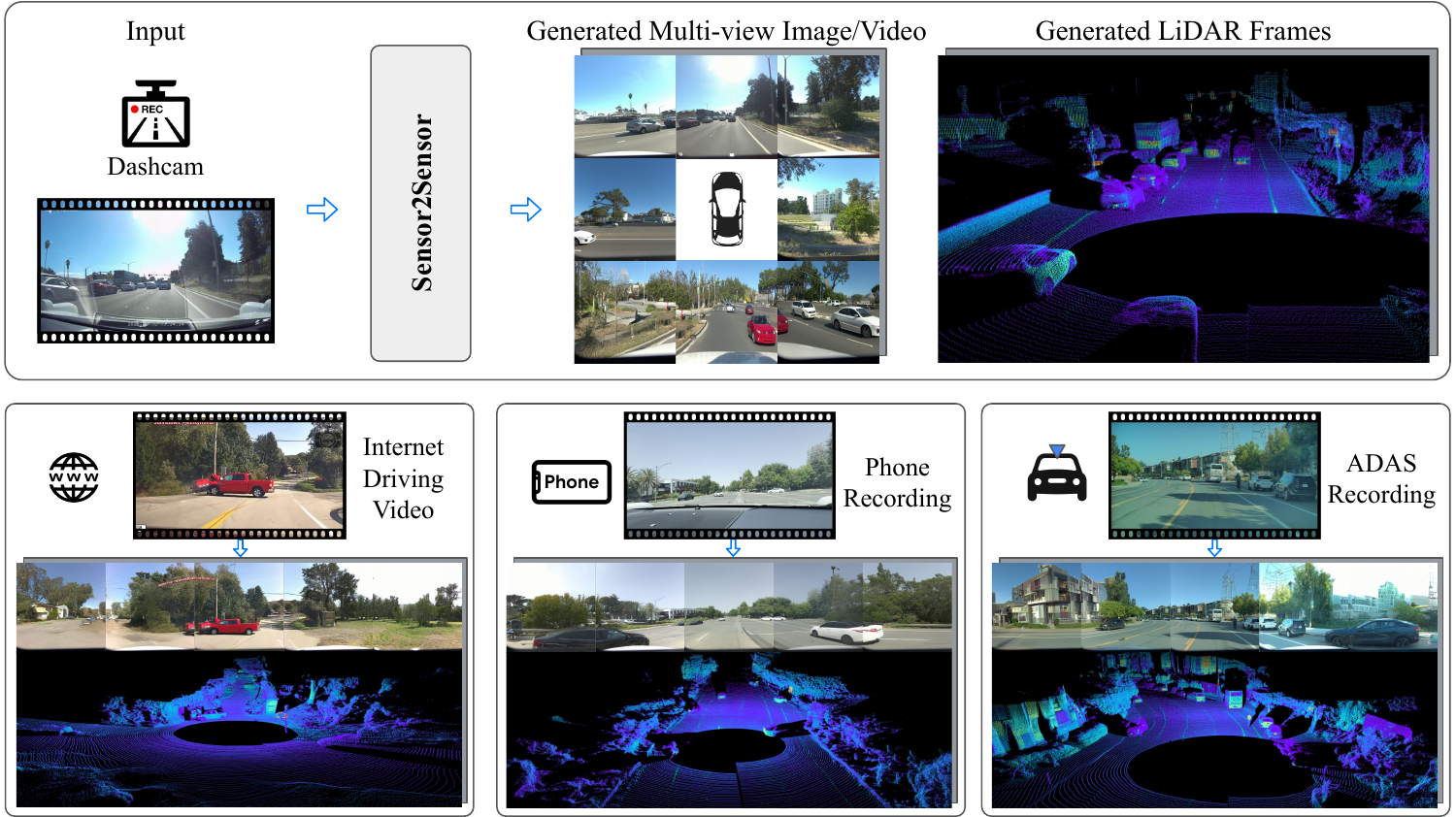} 
    
    \vspace{2mm}
    \captionof{figure}{\textbf{\ours} is a novel generative paradigm for translating in-the-wild monocular videos from varied sources such as dashcams, internet driving videos, phones, and even other Autonomous Driving Systems (ADS), Advanced Driver-Assistance Systems (ADAS) and vehicle platforms into high-fidelity, multi-modal, multi-sensor Autonomous Vehicle (AV) logs specific to a target vehicle embodiment. This enables cross-embodiment sensor conversion to maximally leverage real-world long-tail data for AV system validation.}
    \label{fig:teaser}
\end{strip}
\vspace{2mm}

\begin{abstract}
Robust training and validation of Autonomous Driving Systems (ADS) require massive, diverse datasets. Proprietary data collected by Autonomous Vehicle (AV) fleets, while high-fidelity, are limited in scale, diversity of sensor configurations, as well as geographic and long-tail-behavioral coverage. In contrast, in-the-wild data from sources like dashcams offers immense scale and diversity, capturing critical long-tail scenarios and novel environments. However, this unstructured, in-the-wild video data is incompatible with ADS expecting structured, multi-modal sensor inputs for validation and training.
To bridge this data gap, we propose \ours, a novel generative modeling paradigm that translates in-the-wild monocular dashcam videos into a high-fidelity, multi-modal sensor suite (AV logs) comprising multi-view camera images and LiDAR point clouds. A core challenge is the lack of paired training data. We address this by converting real AV logs into dashcam-style videos via 4D Gaussian Splatting (4DGS) reconstruction and novel-view rendering. \ours~then utilizes a diffusion architecture to perform the generative conversion. We perform comprehensive quantitative evaluations on the fidelity and realism of the generated sensor data. We demonstrate \ours's practical utility by converting challenging in-the-wild internet and dashcam footage into realistic, multi-modal data formats, further unlocking vast external data sources for AV development.
\end{abstract}

\section{Introduction}
\label{sec:intro}

The validation of Autonomous Driving Systems (ADS) against the full spectrum of real-world driving scenarios remains a paramount challenge in the field~\cite{Chen2023EndToEnd}. While generalist policies trained on aggregated data from diverse embodiments have shown promise, they do not obviate the need for rigorous, per-embodiment evaluation. This evaluation is non-negotiable for safety-critical systems, and its efficacy is fundamentally constrained by the profound scarcity of \emph{long-tail data}~\cite{Hegde2025Distilling,Pan2024VLP,zhu2025scenecrafter}.
These long-tail scenarios encompass statistically rare yet safety-critical events, including erratic driving, sudden pedestrian maneuvers, and extreme weather or environmental conditions. Collecting such data organically is prohibitively expensive, requiring fleet-scale operations of immense cost and duration~\cite{Chen2023EndToEnd}.

Two main avenues have been explored to address this data deficiency. The first is \textit{de novo} scenario synthesis using generative models~\cite{Ji2025Txt2Sce, Cai2025Text2Scenario}. While this can create novel events, the generated data often suffers from a critical plausibility gap (non-physical dynamics) and a realism problem (low sensor fidelity) unsuitable for ADS validation.

The second avenue seeks to leverage the immense scale and diversity of ``in-the-wild" third-party data, sourced from internet videos or partner dashcam fleets (Original Equipment Manufacturers, OEMs)~\cite{Miao2024FromDashcam}. These data are, by construction, grounded in physical reality, thus eliminating concerns of event plausibility. It is also naturally skewed towards the long-tail, as mundane events are less likely to be recorded or shared. This approach, however, suffers from a severe \emph{embodiment gap}~\cite{Gao2025Foundation}. This in-the-wild data is sensorially and geometrically misaligned with the target ADS platforms: it typically consists of a single monocular video, lacks the 360-degree multi-camera perspectives, and is devoid of critical modalities like LiDAR. This frames the problem as a highly complex, unpaired domain translation task. Unfortunately, classical unpaired translation methods are ill-equipped to bridge such a vast domain gap, as they lack the strong geometric priors and modal capacity to generate a coherent, temporally-consistent, multi-modal sensor suite from a single, uncalibrated video stream~\cite{Fu2019Geometry}.

In this work, we propose \emph{\ours}, a novel generative paradigm for cross-embodiment sensor conversion that synthesizes the advantages of both paths. As shown in Figure~\ref{fig:teaser}, \ours~inherits the real-world plausibility of in-the-wild data while generatively re-rendering it into the precise, multi-modal format of a target AV embodiment.

The central challenge in training \ours~is the absence of large-scale, paired (dashcam, AV log) training data. We circumvent this limitation by proposing a novel synthetic data-pairing pipeline. We leverage existing %
AV logs, which, by design, contain rich 3D information and 360-degree coverage. This high-fidelity data enables us to first reconstruct a 4D scene representation via dynamic 3D Gaussian Splatting (3DGS)~\cite{Kerbl2023GaussianSplatting, wu20244d}. From this reconstructed scene, we can render novel, synthetic-yet-realistic dashcam views, complete with augmentations of intrinsic and extrinsic parameters sampled from real-world dashcam distributions. This process yields the required paired training corpus: (synthetic dashcam, real AV log).

With this paired dataset, we design \ours~as a conditional diffusion model for multi-sensor (eight cameras) and multi-modal (camera and LiDAR) output, conditioned on the input dashcam video. This use of diffusion for geometrically-aware domain adaptation aligns with recent successes in cross-domain transfer~\cite{Ho2020Denoising, Zhan2024Conditional, Samak2025Sim2Real}.

We validate \ours~through a comprehensive evaluation strategy. Quantitative fidelity is assessed using a bespoke, manually-collected ground-truth dataset. Concurrently, a broad qualitative analysis demonstrates the model's efficacy in converting challenging, real-world in-the-wild videos into realistic and usable sensor logs. Our results affirm that \ours~achieves state-of-the-art (SOTA) fidelity, further unlocking vast, previously-incompatible data sources for AV development.

In summary, our contributions are:
\begin{itemize}
    \item We introduce \ours, a novel generative paradigm for translating in-the-wild monocular videos into high-fidelity, multi-modal, and multi-sensor AV logs specific to a target vehicle embodiment.
    \item We propose a pipeline using dynamic 3D Gaussian Splatting to reconstruct scenes from raw AV logs, rendering paired realistic dashcam views as high-quality training data for diffusion models.
    \item We develop a conditional diffusion architecture, designed to be multi-sensor multi-modal, capable of geometrically-aware cross-embodiment sensor conversion.
    \item We demonstrate, through comprehensive evaluation, that our method further unlocks the vast scale and diversity of in-the-wild video, converting challenging internet footage into realistic, usable data for AV development.
\end{itemize}
\section{Related Works}\label{sec:related}

\noindent\textbf{Generative World Models and High-Fidelity Sensor Synthesis.}
Generative World Models~\cite{bruce2024genie,ha2018world,hafner2019learning,hafner2025training,lu2024genex,wan2025wan,wang2025evoworld,waymo2026worldmodel,assran2025v}, often built upon diffusion architectures~\cite{Ho2020Denoising,peebles2023scalable}, are now foundational for physical AI, enabling the synthesis of photorealistic, physics-based data~\cite{lecun2022path,kim2026cosmos}. %
Prominent examples, such as Wayve's GAIA-1~\cite{GAIA1} and the NVIDIA Cosmos~\cite{Cosmos} platform, primarily target scenario generation, future prediction, and planning for closed-loop simulation~\cite{zhang2025world}. While powerful, their objective is orthogonal to our goal of data \emph{conversion}. However, the success of conditional diffusion in \emph{intra}-embodiment sensor translation validates its use for our complex, multi-modal task. Specifically, Camera-to-LiDAR generation using models like LiDMs~\cite{LiDMs} successfully navigates the spatial and modal mismatch between camera views and 3D point clouds. More recent cross-modality frameworks~\cite{guo2025genesis,singh2024genmm,tang2025omnigen,li2025uniscene,ren2025cosmos} like X-Drive~\cite{xie2025xdrive} further demonstrate the ability to generate consistent multi-sensor data. \ours~extends this conditional diffusion capability to the more challenging \emph{cross-embodiment} setting, translating a single monocular stream into a geometrically-accurate, multi-sensor AV log. This complex translation necessitates a geometrically-anchored training corpus, which motivates our integration of reconstructive techniques.

\noindent\textbf{Reconstructive World Models and 4D Scene Representation.}
Reconstructive World Models are essential for high-fidelity 4D (spatio-temporal) scene representation~\cite{song2025coda,peng2025desire,huang2024textit,wang2024dc,ren2024scube,wang2025flux4d}, enabling closed-loop evaluation and novel view synthesis~\cite{DriveDreamer4D}. Advances in explicit representations~\cite{wu20244d}, particularly 3D Gaussian Splatting (3DGS)~\cite{Kerbl2023GaussianSplatting}, have allowed for real-time, photorealistic rendering and dynamic scene modeling in autonomous driving~\cite{PAGS}. Methods like PAGS~\cite{PAGS} and Driv3R~\cite{Driv3R} focus on decomposing the scene or achieving fast, dense 4D reconstruction from multi-view inputs, ensuring geometric accuracy and temporal consistency. These models serve as powerful ``data machines" to augment viewpoints, as seen in works like DriveDreamer4D~\cite{DriveDreamer4D}. \ours~critically repurposes this reconstructive capability to resolve the training data bottleneck~\cite{wang2025drive}. We reconstruct scenes from existing AV logs via 4DGS, treating the reconstruction as a geometric oracle. This allows us to render a synthetic dashcam view from a novel, external viewpoint~\cite{DriveDreamer4D}. This process yields a perfectly paired training corpus, transforming the cross-embodiment challenge into a fully supervised, geometrically-anchored generation task.
\section{Method}
\label{sec:method}

Our approach consists of two key stages: (1) a scalable data curation pipeline using 4DGS to synthesize paired training data (Section~\ref{sec:method_data}), and (2) a diffusion model that generates synchronized multi-view imagery and LiDAR point clouds conditioned on a single camera input (Section~\ref{sec:method_model}). We further extend this to temporally consistent video generation via auto-regressive modeling (Section~\ref{sec:method_video}).

\subsection{Synthetic Sensor Simulation via 4DGS}
\label{sec:method_data}
\begin{figure}%
    \centering
    \includegraphics[width=\columnwidth]{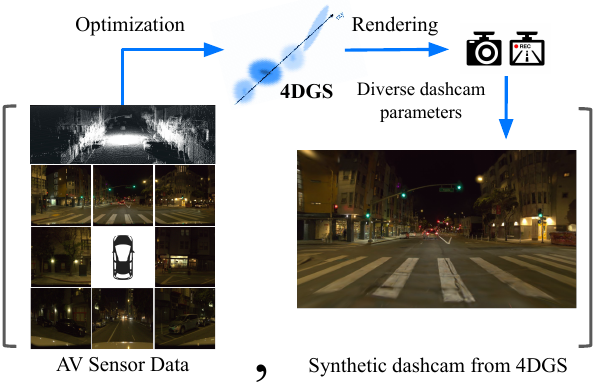}
    \caption{\textbf{Synthetic paired-data curation pipeline.} We reconstruct 4DGS from 8-view cameras and render a diverse set of synthetic third-party cameras (e.g. popular dashcam models).}
    \label{fig:datapipeline}
    \vspace{-1em}
\end{figure}

\noindent\textbf{4DGS for Autonomous Driving.}
We use a variant of 3D Gaussian Splatting (3DGS) \cite{Kerbl2023GaussianSplatting} with support for dynamic rigid (e.g. vehicles) and deformable (e.g. pedestrian) objects to construct 4D representations  of diverse AV scenarios. In total, approximately 100,000 scenes of 10s duration were chosen for reconstruction. Each scene contains multi-view camera data spanning 360 degrees as well as LiDAR data, which is used to initialize and regularize the geometry of the 3D Gaussian Splats, though optional. Splats belonging to moving objects are accumulated using a canonical object model to achieve more complete object coverage. Once a scene is optimized, it can be rendered using virtual cameras with augmented intrinsic and extrinsic parameters to mimic the optics and placement of dashcams found in-the-wild. Note that due to the purely reconstructive nature of 3DGS, the best rendering quality is achieved within a bounded region around the original camera poses. Unlike the original 3DGS formulation, we use a ray-tracing-based rendering approach to better support fish-eye optics.

\begin{figure*}%
    \centering
    \includegraphics[width=1.9\columnwidth]{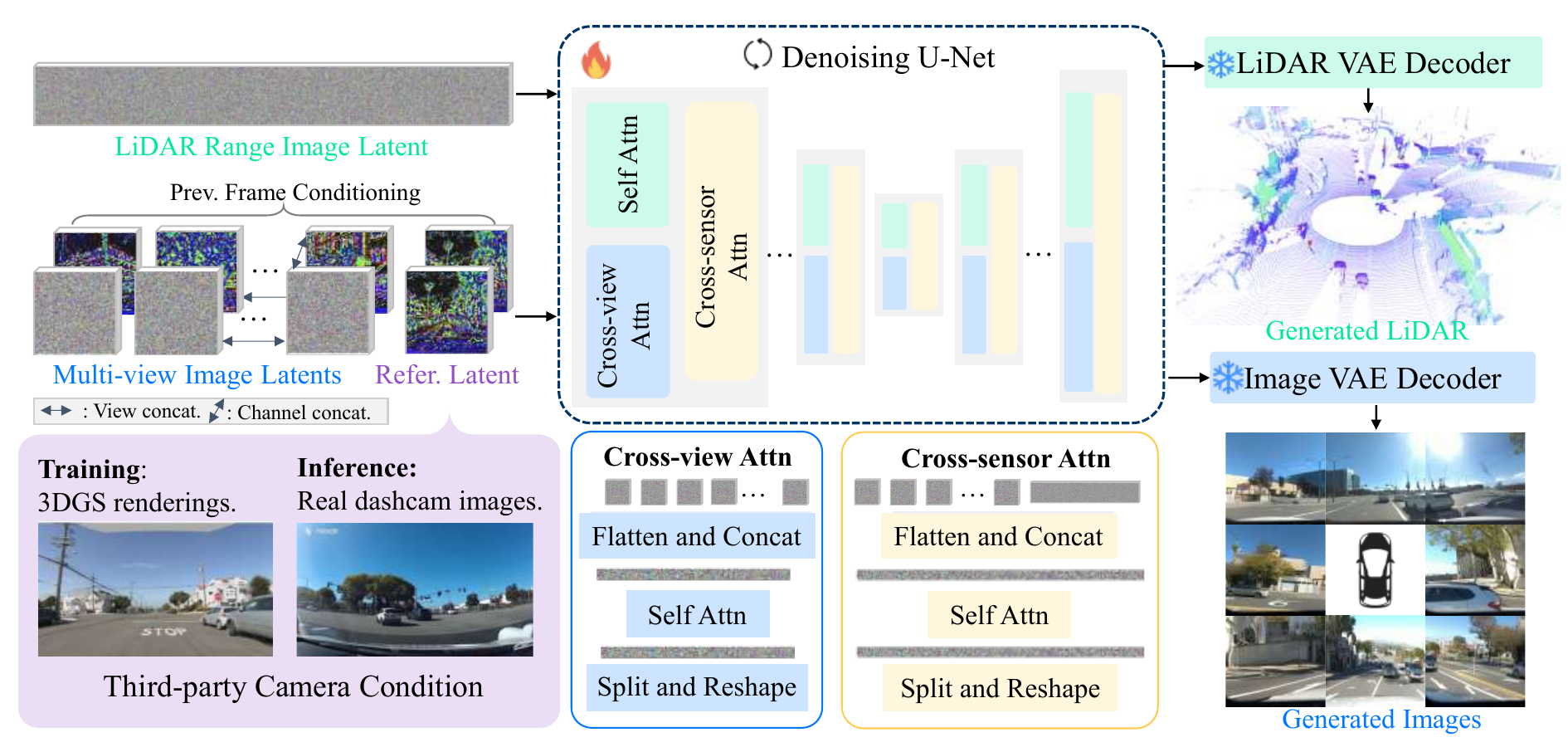}
    \caption{\textbf{Our multi-modal, multi-view sensor generation model architecture.} Based on Latent Diffusion, the model simultaneously generates multi-view images ($C$) and LiDAR point clouds ($L$) using modality-specific VAEs and U-Net towers. \textbf{Multi-sensor consistency} is enforced via cross-sensor attention, and \textbf{multi-view consistency} is maintained with 3D attention blocks.}
    \label{fig:arch}
    \vspace{-1em}
\end{figure*}

\noindent\textbf{Third-party Camera Synthesis.}
We leverage high-fidelity 4DGS representations to synthesize a large, paired training corpus by rendering virtual cameras (Figure~\ref{fig:datapipeline}). This process explicitly bridges the domain gap between the source sensor data and the target third-party sensors (e.g., dashcams). The synthesis pipeline models two primary sources of sensor variation found in off-the-shelf dashcam systems: \emph{Intrinsic Parameters} ($\mathbf{p}_i$): Generated by sampling realistic focal lengths, principal points, and distortion coefficients ($\mathbf{\kappa}$). This step emulates the diverse optical profiles of low-cost, wide-angle lenses prone to significant distortion. \emph{Extrinsic Parameters} ($\mathbf{p}_e$): Sampled as 6-DoF poses, $\mathbf{p}_e = [\mathbf{R} | \mathbf{t}]$, relative to the vehicle frame. This accounts for variations in vehicle type, diverse mounting locations (e.g., driver-side), and minor rotational perturbations ($\theta_p, \theta_y, \theta_r$) simulating imperfect camera installation. This rendering approach creates a vast dataset where each dashcam-style frame is perfectly time-synchronized and spatially aligned with the ground truth sensors.

\subsection{Multi-modal Diffusion Model for Sensors}
\label{sec:method_model}
To enable sensor conversion from third-party data, we first develop a \textbf{multi-sensor, multi-view} generation model. This model simultaneously generates multi-view images $C=\{\mathbf{c}_i\}_{i=1}^N$ and the LiDAR point cloud $L$. Each sensor modality has its own VAE and U-Net branch for diffusion. The key attributes of this model are multi-view (Section~\ref{sec:multi-view}) and multi-sensor (Section~\ref{sec:multi-sensor}) consistency.

\subsubsection{Multi-view Image Generation} 
\label{sec:multi-view}
The image branch builds on a multi-view diffusion model~\cite{CAT3D} that enables view consistency and camera pose control over the image generation. Given the camera parameters for each camera, this model learns a joint distribution of all images. 
To achieve multi-view consistency, the model replaces the $2$D attention modules in the original LDM to $3$D ($1$D cross views and $2$D in spatial) and computes attentions on all images. 

Furthermore, to precisely control the poses of generated images, this model accepts camera parameters as conditions. The camera parameters are represented via raymaps~\cite{sitzmann2021light, CAT3D}, which encode the ray origin and direction at each spatial location. All raymaps are normalized with regard to the first camera and %
concatenated channel-wise onto the image features.  

\subsubsection{LiDAR Generation}

\noindent \textbf{LiDAR Representation.}
To effectively leverage the capabilities of 2D generative models, we utilize the LiDAR point cloud's native representation as range-view spin images---a tensor with shape $[H_L, W_L, D_L]$, where the $D_L=4$ channels correspond to (1) range (depth in meters), (2) intensity (amount of light reflected), (3) elongation (to what extent the waveform has been ``flattened"), and (4) validity (1 for a return, 0 otherwise). 
The image rows and columns map to the sensor's elevation and azimuth angles, respectively. Each (row, col, range) value can be projected to and from 3D Euclidean space $(x, y, z)$ given the vehicle trajectory and sensor calibration. For normalization, range values are clamped at 150 meters and linearly scaled to the $[0, 1]$ interval. Intensity and elongation are similarly normalized to fit within $[0, 1]$.

\noindent \textbf{LiDAR VAE.}
We introduce a VAE architecture for generating LiDAR spin images, jointly encoding depth, intensity, and elongation. The encoder and decoder are both convolutional, %
and we optimize the VAE via 
\begin{align}
\label{eq:lidar_loss_total}
\mathcal{L}^{\text{TOTAL}} = 
&\mathcal{L}^{\text{L1}}_{\text{range}} + \mathcal{L}^{\text{L1}}_{\text{elongation}} + \mathcal{L}^{\text{L1}}_{\text{intensity}} + \mathcal{L}^{\text{BCE}}_{\text{validity}} + \mathcal{L}^{\text{LPIPS}}_{\text{normals}} \nonumber \\
& + \mathcal{L}^{\text{LPIPS}}_{\text{elongation}} + \mathcal{L}^{\text{LPIPS}}_{\text{intensity}} + \mathcal{L}^{\text{LPIPS}}_{\text{validity}} + \mathcal{L}^{\text{KL}}.
\end{align}
Additional training details are provided in the supplemental.

\noindent \textbf{LiDAR Diffusion.}
We first project the raw LiDAR range images into a latent space using the LiDAR VAE. A LiDAR U-Net branch then performs diffusion on this latent, operating similarly to a standard single-view image diffusion model. Each layer in the LiDAR U-Net is designed to output a feature with the same channel dimension as its corresponding layer in the multi-view image branch, enabling our cross-sensor feature fusion.

\subsubsection{Cross-Sensor Attention Module}
\label{sec:multi-sensor}
As shown in Figure~\ref{fig:arch}, to simultaneously generate consistent images and LiDAR, we introduce a cross-sensor attention module within each U-Net block. We inject this module after convolutional layers to promote continuous information interchange. In detail, at a given block $i$, we flatten the image features $\mathbf{f}_C^i$ and LiDAR features $\mathbf{f}_L^i$ into token sequences $\mathbf{T}_C^i \in \mathbb{R}^{K_C \times d^i}$ and $\mathbf{T}_L^i \in \mathbb{R}^{K_L \times d^i}$, where $K_C = N \times h_C^i \times w_C^i$ and $K_L = h_L^i \times w_L^i$. The shared U-Net architecture for both modalities ensures their feature dimension $d^i$ is identical. These tokens are then concatenated into a unified sequence $\mathbf{T}_U^i \in \mathbb{R}^{(K_C + K_L) \times d^i}$, and the module computes self-attention over this sequence, allowing features from both sensors to interact directly.

\subsubsection{Third-party Camera Condition}
To directly leverage the visual context of the third-party data (e.g., dashcams), we introduce it as an additional, conditional ninth view, distinct from the $N=8$ views targeted for generation. This conditional input is processed by the encoder to generate a latent representation, which is then concatenated with (1) a corresponding raymap~\cite{sitzmann2021light, CAT3D} and (2) a binary conditioning mask. This mask explicitly signals to the model that this view is a known, noise-free condition, distinguishing it from the $N$ noisy latents to be denoised. This augmented latent is then concatenated along the view dimension with the latents from the original eight views, and the resulting $(N+1) \times H \times W \times C$ tensor is processed by the diffusion layers. This allows the features from the $8$ target views to interact with the conditional view through attention, effectively conditioning the synthesis of the surrounding scene on the dashcam's context. This $9^{th}$ view is excluded from the loss computation, ensuring its role as a conditioning input and that the network's capacity is focused on accurately generating the eight target views.

\subsection{Auto-regressive Video Generation}
\label{sec:method_video}

To convert third-party videos to driving logs, we extend our model for auto-regressive generation. Given the third-party camera frame $\mathbf{x}_t$ at time step $t>0$, we aim to model the conditional probability distribution of the multi-view images $C_t$ and LiDAR point cloud $L_t$, conditioning on the self-generations at step $t-1$:
\begin{equation}
  P(C_t, L_t | \mathbf{x}_t, C_{t-1}, L_{t-1}).
\end{equation}
When $t=0$, sensor data is generated conditioning only on $\mathbf{x}_0$. Vanilla auto-regressive generation suffers from drifting, as models trained on ground-truth (GT) context must generate sequences conditioned on their own imperfect generations during inference. This causes errors to accumulate over long rollouts. To mitigate this, we introduce the \textit{DAgger} algorithm~\cite{DAGGER}, which augments the training context with the model's own generations. We gradually shrink this train-test mismatch by iteratively generating rollout videos and training a new model on the resulting context. To maintain robustness, we set a 0.2 probability of training on the original GT context.

\section{Experiments}
\label{sec:exp}
\begin{figure*}%
    \centering
    \includegraphics[width=\linewidth]{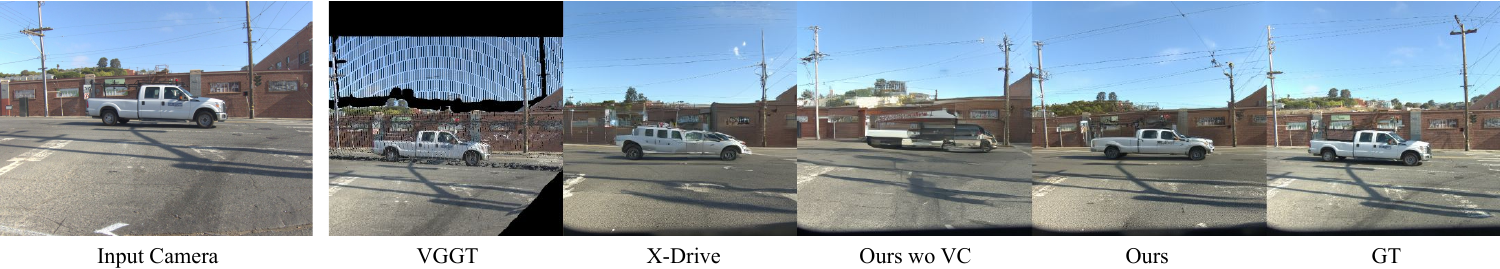}
    \caption{\textbf{Image comparison}. Our method \ours~produces results largely faithful to the ground truth, while the baselines either fail to preserve the scene and object structures, or cannot create plausible generations of the unobserved areas.}
    \label{fig:image_comparison}
\end{figure*}
Our experiments are designed to: (1) quantify the fidelity of our generated images, video, and LiDAR point clouds against strong baselines; (2) test model's generalizability on challenging, in-the-wild driving footage; and (3) validate key architectural and training choices via ablation studies.

\subsection{Experiment Settings}
\noindent\textbf{Evaluation metrics.} We evaluate our results using Fr\'echet Inception Distance (FID) ($\downarrow$) \cite{heusel2017gans}  for image realism and Fr\'echet Video Distance (FVD) ($\downarrow$) \cite{unterthiner2019fvd} for video realism. For paired ground-truth comparisons, we use Peak Signal-to-Noise Ratio (PSNR) ($\uparrow$), Structural Similarity Index Measure (SSIM) ($\uparrow$) \cite{wang2004image}, and the Learned Perceptual Image Patch Similarity (LPIPS) ($\downarrow$) \cite{zhang2018perceptual}. These are supplemented by Human Evaluation ($\uparrow$), where raters choose the more realistic result in side-by-side comparisons.

\noindent\textbf{Dataset.} Since paired, third-party-to-AV sensor generation is a novel task, no public datasets with such synchronized data exist for evaluation. We therefore curated an evaluation dataset comprising two key components: (a) A dataset of 1,000 paired ``Fixed-Camera-to-AV" log sequences (each ~3 seconds long). The fixed-camera is a bumper camera positioned at the front-left bumper of the AV, and the 8-view surrounding cameras and the LiDAR are on top of the AV. (b) An ``in-the-wild" dataset, including manually-collected real dashcam recordings, driving videos available on the internet, phone recordings and footage from other ADAS, for showing the in-the-wild generalizability.

\noindent\textbf{Baselines.} 
End-to-end conversion of a monocular third-party video to a full AV sensor suite (multi-view cameras and LiDAR) has not been fully explored in previous work. Thus, no direct baselines exist for our specific task. To benchmark \ours, we adapted several state-of-the-art  methods for comparison.
Reconstruction-based: We compare against state-of-the-art feedforward 3D scene reconstruction models VGGT \cite{wang2025vggt} and $\pi^3$ \cite{wang2025pi3} for the multi-camera generation task. 
Generative models: We adapt two SOTA generative models. X-Drive \cite{xie2025xdrive}, an image-LiDAR co-generation model, was modified to condition on the dashcam input via attention. We also adapted CAT3D \cite{CAT3D} by (1) enabling LiDAR generation using the same VAE as our method and (2) conditioning it on the dashcam via channel-concatenation (CC) instead of view-concatenation (VC). We refer to this baseline as ``Ours without (wo) VC", which also serves as a key ablation against our approach.

\subsection{Multi-view Image Generation}
We first evaluate the task of multi-view image generation. To quantitatively measure performance, we curate a ``Fixed-Camera-to-AV'' dataset. The input for this task comes from a real, front-left facing camera fixed on the AV near the bumper. This input camera is synchronized and calibrated with the target 8 surrounding views, to provide an accurate quantitative benchmark, as shown in Table~\ref{tab:image_eval}.

\begin{table}[t!]
\centering
\caption{\textbf{Evaluation on multi-view image generation from a fixed bumper camera}. We compare our method against baselines on our paired dataset. $\downarrow$: Lower is better. $\uparrow$: Higher is better. VC means concatenating dashcam in the view dimension.}
\label{tab:image_eval}
\setlength{\tabcolsep}{10pt}          %
\resizebox{0.47\textwidth}{!}{%
\begin{tabular}{l cccc}
\toprule
Method & FID $\downarrow$ & PSNR $\uparrow$ & SSIM $\uparrow$ & LPIPS $\downarrow$ \\
\midrule
VGGT \cite{wang2025vggt} & 250.93 & 14.73 & 0.433 & 0.491 \\
$\pi^3$ \cite{wang2025pi3} & 246.27 & 14.93 & 0.470 & 0.458 \\
X-Drive \cite{xie2025xdrive} & 8.30 & 18.61 & 0.536 & 0.345 \\
Ours without VC  & 6.88 & 18.69 & 0.531 & 0.346 \\
\textbf{Ours} & \textbf{6.47} & \textbf{19.06} & \textbf{0.539} & \textbf{0.316} \\
\bottomrule
\end{tabular}%
}
\end{table}

On this ``Fixed-Camera-to-AV'' generation task, our method outperforms all baselines with an FID of \textbf{6.47} and LPIPS of \textbf{0.316}, demonstrating the superior generative quality.
Figure~\ref{fig:image_comparison} shows that images generated by \ours~are clear, geometrically plausible, and maintain consistent appearance of objects as they appear between camera views. In contrast, baseline methods often produce blurry results, distorted geometry, or noticeable artifacts.

\subsection{Video Generation}
\begin{figure}%
    \centering
    \includegraphics[width=\columnwidth]{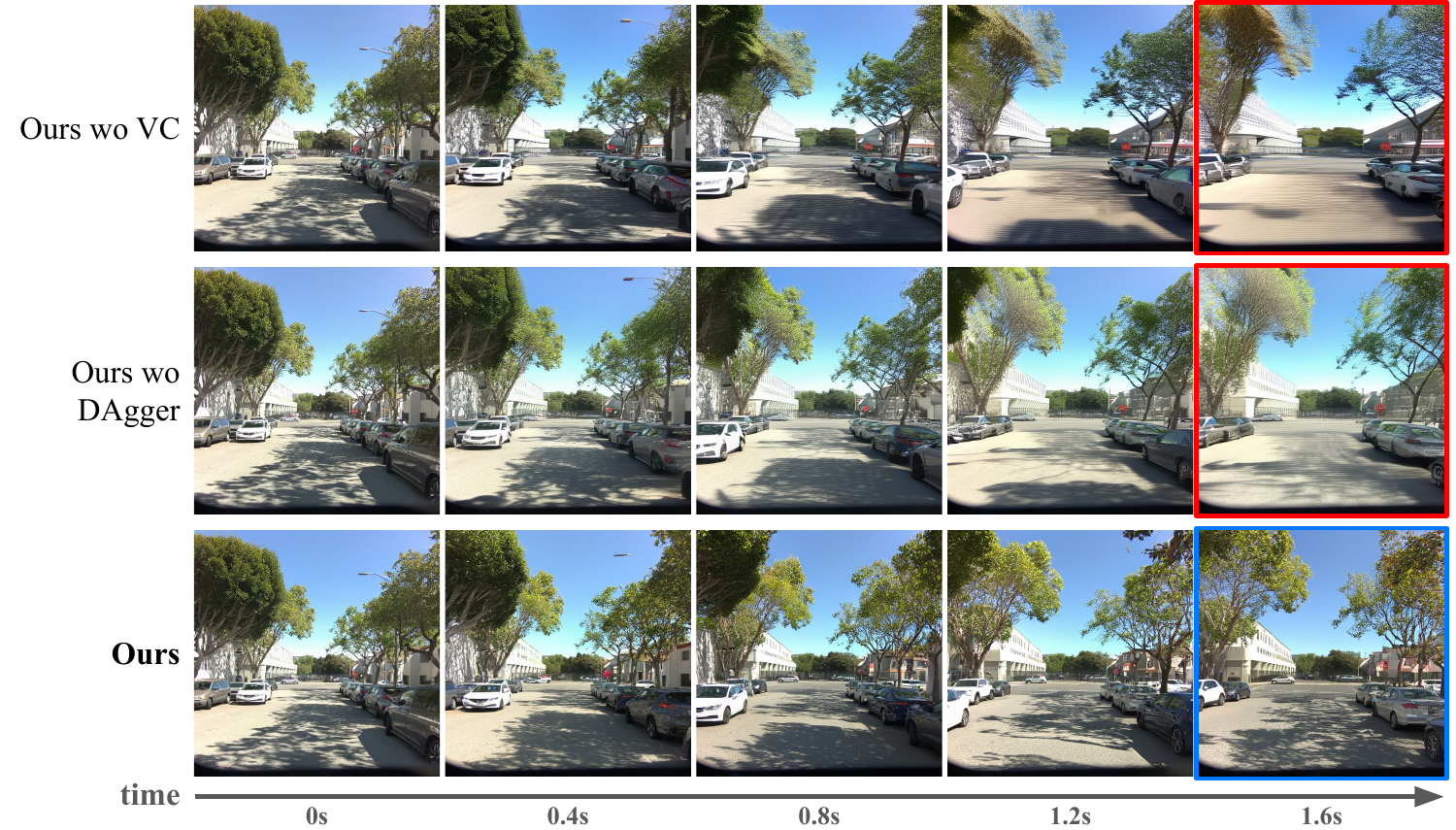}
    \caption{\textbf{Temporal video rollout comparison} (only showing front view for compactness). DAgger training significantly improves temporal stability of generated videos through the rollout.}
    \label{fig:video}
\end{figure}
\begin{figure*}%
    \centering
    \includegraphics[width=\linewidth]{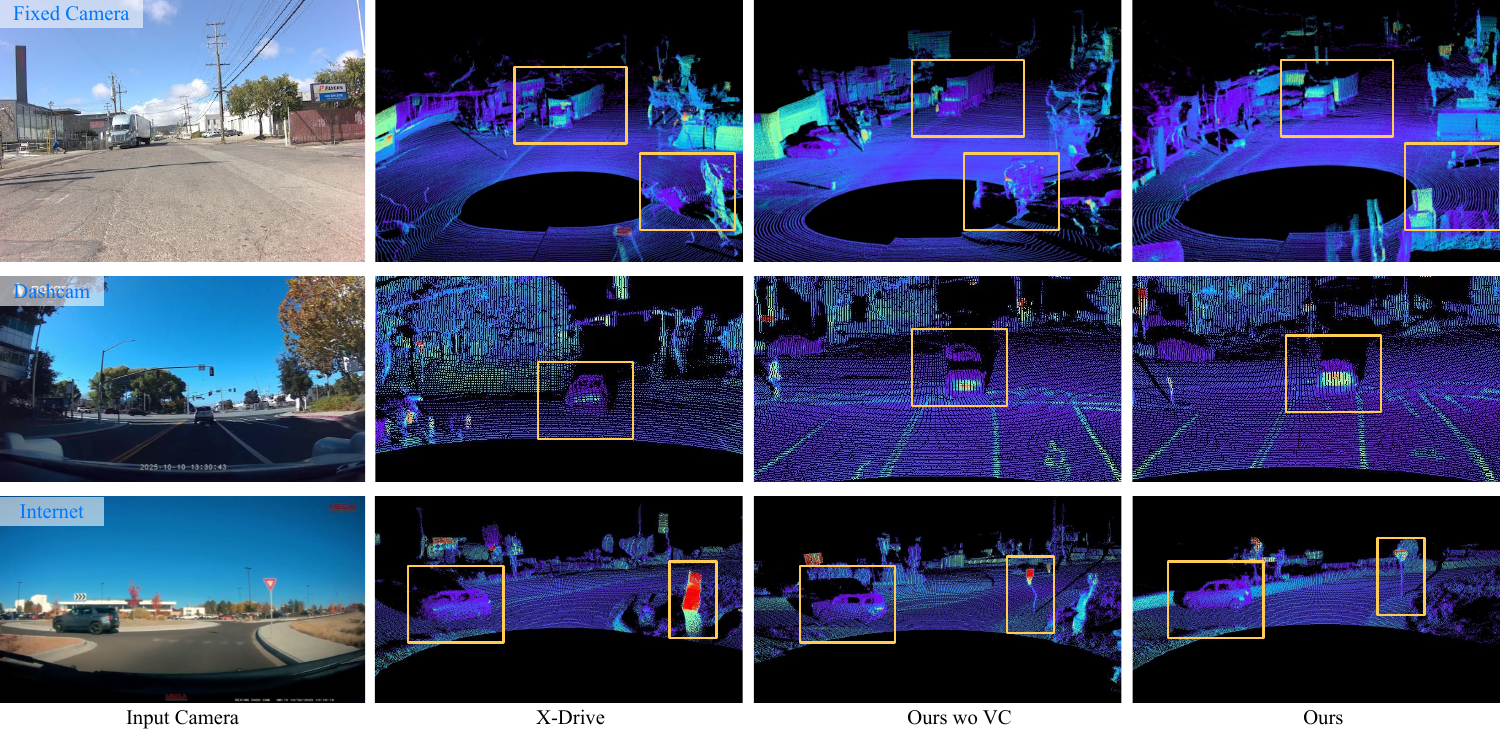}
    \caption{\textbf{Qualitative LiDAR Comparison}. Our method correctly renders the truck's shape and has less noise in the surrounding objects, while the other methods produce distortions and incorrect intensity. All methods use the same LiDAR VAE for a fair comparison.}
    \label{fig:lidar}
\end{figure*}

Beyond static images, we evaluate the temporal consistency of our generated multi-view videos. We report quantitative results on our paired ``Fixed-Camera-to-AV'' dataset in Table~\ref{tab:video_eval}. We use Fr\'echet Video Distance (FVD) ($\downarrow$) as the primary metric for overall video quality, supplemented by frame-wise PSNR ($\uparrow$), SSIM ($\uparrow$), and LPIPS ($\downarrow$). X-Drive is excluded from this comparison, as it is an image-only model and does not generate video. Furthermore, the reconstruction-based methods (VGGT and $\pi^3$) only generate complete results for the front view, as their other views suffer from large empty regions. For a better comparison, all metrics in this table are computed exclusively on the generated front-view videos.

Our model shows superior temporal stability, achieving the best FVD of 278.12. This significantly outperforms all baselines, such as Ours wo VC (293.73) and the feedforward reconstruction models $\pi^3$ (2007.35) and VGGT (2373.15). The feedforward models' high FVD scores are expected, as their reconstructive-only design cannot produce coherent novel views. This indicates that we not only generate realistic individual frames but also ensure they are coherent over time. The strong per-frame metrics (PSNR 22.42, SSIM 0.623, LPIPS 0.186) further support this, reinforcing the high fidelity seen in our static image evaluation.

 Moreover, as shown in~Figure~\ref{fig:video}, while baselines exhibit noticeable flickering or inconsistent object appearance across frames, our model produces smooth and coherent video sequences for all views, which is critical for downstream consumption by perception or simulation systems.

\begin{table}[t!]
\centering
\caption{\textbf{Evaluation on multi-view video generation}. We compare on the ``Fixed-Camera-to-AV" dataset.}
\label{tab:video_eval}
\setlength{\tabcolsep}{10pt}          %
\resizebox{\columnwidth}{!}{%
\begin{tabular}{l cccc}
\toprule
Method & FVD $\downarrow$ & PSNR $\uparrow$ & SSIM $\uparrow$ & LPIPS $\downarrow$ \\
\midrule
VGGT \cite{wang2025vggt} & 2373.15 & 14.73 & 0.433 & 0.491 \\
$\pi^3$ \cite{wang2025pi3} & 2007.35 & 14.93 & 0.470 & 0.458  \\
Ours without VC & 293.73 & 22.07 & 0.622 & 0.204 \\
\textbf{Ours} & \textbf{278.12} & \textbf{22.42}  & \textbf{0.623} & \textbf{0.186} \\
\bottomrule
\end{tabular}%
}
\end{table}

\subsection{LiDAR Generation}

\begin{figure}%
    \centering
    \includegraphics[width=\columnwidth]{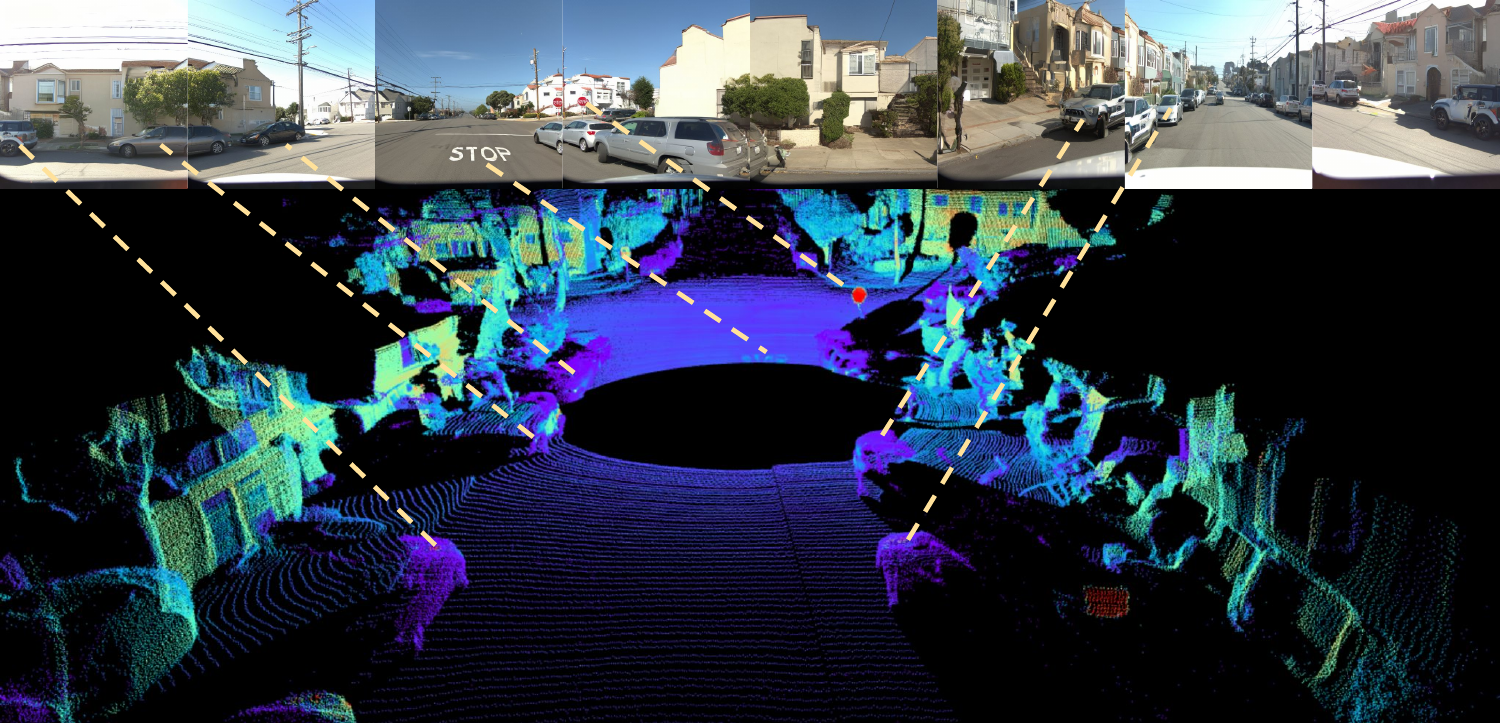}
    \caption{\textbf{Visualization of joint image and LiDAR generation}. \ours~achieves cross-modal consistency between image and LiDAR,  faithfully generating safety-critical objects, including signage, road markings, and vehicles.}
    \label{fig:joint}
\end{figure}

A key contribution of \ours~is its multi-modal capability to co-generate LiDAR point clouds along with multi-view videos.
Qualitatively, Figure~\ref{fig:lidar} provides a direct comparison against baseline methods. Our model shows a superior ability to reconstruct plausible 3D geometry for both nearby actors (like the truck) and the static environment. Our results are cleaner, with fewer noise artifacts and more accurate intensity rendering compared to X-Drive and Ours wo VC.
Furthermore, Figure~\ref{fig:joint} highlights our model's strength in producing \emph{jointly consistent} image and LiDAR outputs. The generated LiDAR points correctly align with their corresponding objects in the generated camera views, demonstrating that the model has learned a coherent underlying 3D representation of the scene.

Quantitatively, we report the Chamfer Distance for generated LiDAR in Table~\ref{tab:lidar_eval}. Moreover, human evaluation of LiDAR generation in Table~\ref{tab:human_eval} also demonstrates a clear preference for our generated LiDAR over the baselines. 

\begin{table}[]
\centering
\caption{\textbf{LiDAR Generation Accuracy.} Comparison of Chamfer Distance ($\downarrow$) between the baseline and our proposed method.}
\label{tab:lidar_eval}
\resizebox{0.80\columnwidth}{!}{%
\begin{tabular}{lcc}
\toprule
\textbf{Method} & \textbf{Chamfer Distance $\downarrow$} & \textbf{Improvement (\%)} \\
\midrule
X-Drive~\cite{xie2025xdrive}         & 10.02                    & ---                  \\
\textbf{Ours}   & \textbf{8.68}           & \textbf{13.37\%}      \\
\bottomrule
\end{tabular}
}
\vspace{-8pt}
\end{table}

\subsection{Generalization on in-the-wild driving data}
\begin{figure*}%
    \centering
    \includegraphics[width=\linewidth]{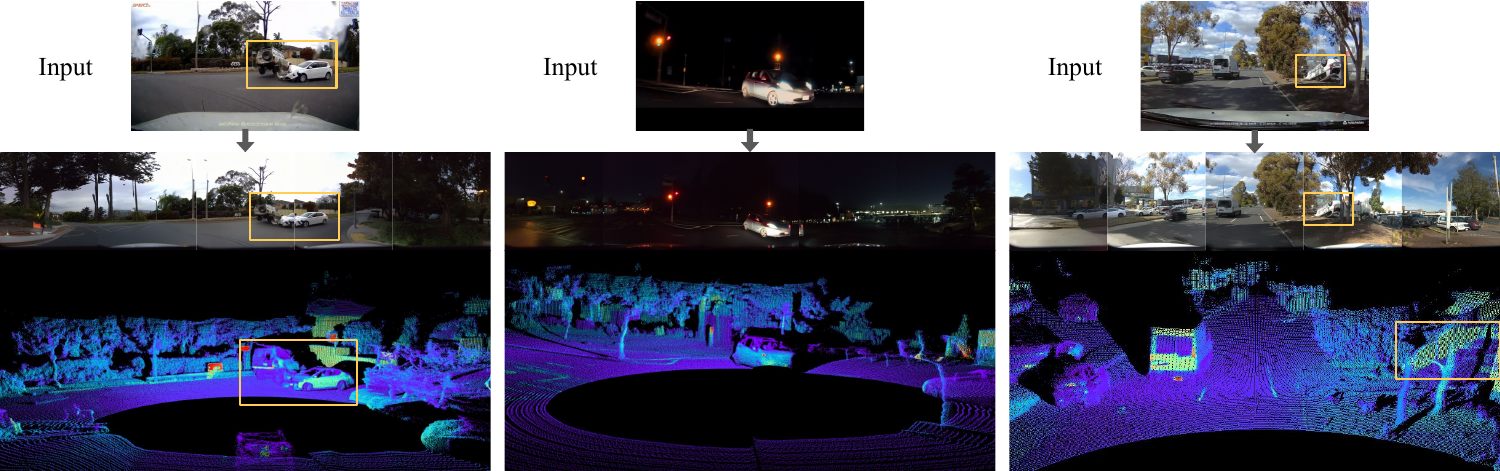}
    \caption{\textbf{Qualitative generalization to in-the-wild internet videos}. \ours~successfully converts diverse and challenging monocular inputs, including long-tail crashes, night-time scenes with low visibility, and active incidents, into full, coherent AV sensor suites.}
    \label{fig:ood}
\end{figure*}

\begin{table}[t!]
\centering
\caption{\textbf{Human evaluation for in-the-wild generation}. We show top-rank (top half) and pair-wise (bottom half) preference rates. Participants were asked to rank results of all three methods based on realism and alignment to the input.}
\label{tab:human_eval}
\resizebox{\columnwidth}{!}{%
\begin{tabular}{l cc cc}
\toprule
& \multicolumn{2}{c}{Dashcam} & \multicolumn{2}{c}{Internet} \\
\cmidrule(lr){2-3} \cmidrule(lr){4-5}
Method & Image $\uparrow$ & LiDAR $\uparrow$ & Image $\uparrow$ & LiDAR $\uparrow$ \\
\midrule
X-Drive \cite{xie2025xdrive} & 3.08\% & 8.08\% & 1.54\% & 7.69\% \\
Ours without VC & 13.46\% & 23.85\% & 13.85\% & 33.85\% \\
\textbf{Ours} & \textbf{83.46\%} & \textbf{68.08\%}  & \textbf{84.62\%} & \textbf{58.46\%}\\
\midrule
Ours without VC $>$ X-Drive & 67.69\% & 69.62\% & 84.62\% & 73.46\% \\
Ours $>$ Ours without VC & 85.77\% & 73.46\% & 85\% & 63.46\% \\
Ours $>$ X-Drive & 94.62\% & 87.31\% & 95.38\% & 85\% \\
\bottomrule
\end{tabular}%
}
\vspace{-1em}
\end{table}

The primary motivation for \ours~is to further unlock ``in-the-wild" data. We test this by applying our model, trained only on our paired dataset, to a diverse set of uncurated videos from internet, dashcams, and other third-party sources. These videos feature camera intrinsics, extrinsics, weather conditions and content unseen during training.

As shown in Fig.~\ref{fig:ood}, \ours demonstrates strong qualitative generalization. Despite facing unknown sensor characteristics and challenging, unseen environments (such as night-time near collisions, accidents, and active incidents), our model converts monocular inputs into coherent multi-sensor AV logs while preserving critical scene elements. This highlights its robustness for mining long-tail scenarios from vast, previously incompatible data sources.

Quantitatively, a comprehensive human evaluation is shown in Table~\ref{tab:human_eval}. %
26 participants evaluated $40\times3$ generated image and LiDAR samples based on realism and alignment with the input image. After training and qualification on the protocol, they ranked each triplet as \textit{best}, \textit{middle}, or \textit{worst}, from which we computed top-rank and pairwise preference rates.
On dashcam data, \ours is top-preferred in 83.46\% of image cases and 68.08\% for LiDAR; on internet data, 84.62\% and 58.46\%, respectively. Pairwise comparisons show \ours is preferred over X-Drive in over \textbf{94\%} of image cases and \textbf{85\%} for LiDAR.

\subsection{Ablation Study}
\begin{table}[t!]
\centering
\caption{\textbf{Ablation on model architecture}. We compare input conditioning (CC vs. VC) and the impact of joint LiDAR training, evaluated on the ``Fixed-Camera-to-AV'' dataset. CC is channel concatenation, VC is view concatenation.}
\label{tab:ablation_arch}
\resizebox{0.48\textwidth}{!}{%
\begin{tabular}{l cccc}
\toprule
Method & FID $\downarrow$ & PSNR $\uparrow$ & SSIM $\uparrow$ & LPIPS $\downarrow$ \\
\midrule
CAT3D + CC (image-only) & 6.63 & 18.91 & 0.542 & 0.314 \\
CAT3D + VC (image-only) & \textbf{6.20} & \textbf{19.12} & \textbf{0.543} & \textbf{0.307} \\
\midrule
CAT3D + CC + LiDAR & 6.88 & 18.69 & 0.531 & 0.346 \\
CAT3D + VC + LiDAR (\textbf{ours}) & \textbf{6.47} & \textbf{19.06} & \textbf{0.539} & \textbf{0.316}  \\
\bottomrule
\end{tabular}%
}
\end{table}

\begin{table}[t!]
\centering
\caption{\textbf{Ablation on DAgger finetuning for video generation.}}
\label{tab:ablation_dagger}
\setlength{\tabcolsep}{10pt}          %
\resizebox{0.48\textwidth}{!}{%
\begin{tabular}{l cc}
\toprule
Method & Front-view FVD $\downarrow$ & Front-view FID $\downarrow$ \\
\midrule
Without DAgger & 288.90  & 24.65 \\
With DAgger (\textbf{ours}) & \textbf{278.12} & \textbf{21.54}  \\
\bottomrule
\end{tabular}%
}
\vspace{-1em}
\end{table}

\noindent\textbf{Model Architecture.} Table~\ref{tab:ablation_arch} analyzes key architectural choices. 
First, we compare input conditioning via channel concatenation (CC) and view concatenation (VC). In the image-only setting, VC achieves better FID (6.20 vs. 6.63).
Second, we study joint image-LiDAR training. Our full model achieves LPIPS 0.316, outperforming the CC variant (0.346) while remaining competitive with image-only VC (0.307). This confirms that our design enables joint LiDAR generation without obvious image quality degradation.

\noindent\textbf{DAgger Finetuning.}
Table~\ref{tab:ablation_dagger} shows that DAgger finetuning improves video quality. With DAgger, FVD and FID improve to 278.12 and 21.54. This demonstrates improved temporal consistency and fidelity.

\subsection{Downstream Tasks}
\begin{figure}[]
\centering
\includegraphics[width=\columnwidth]{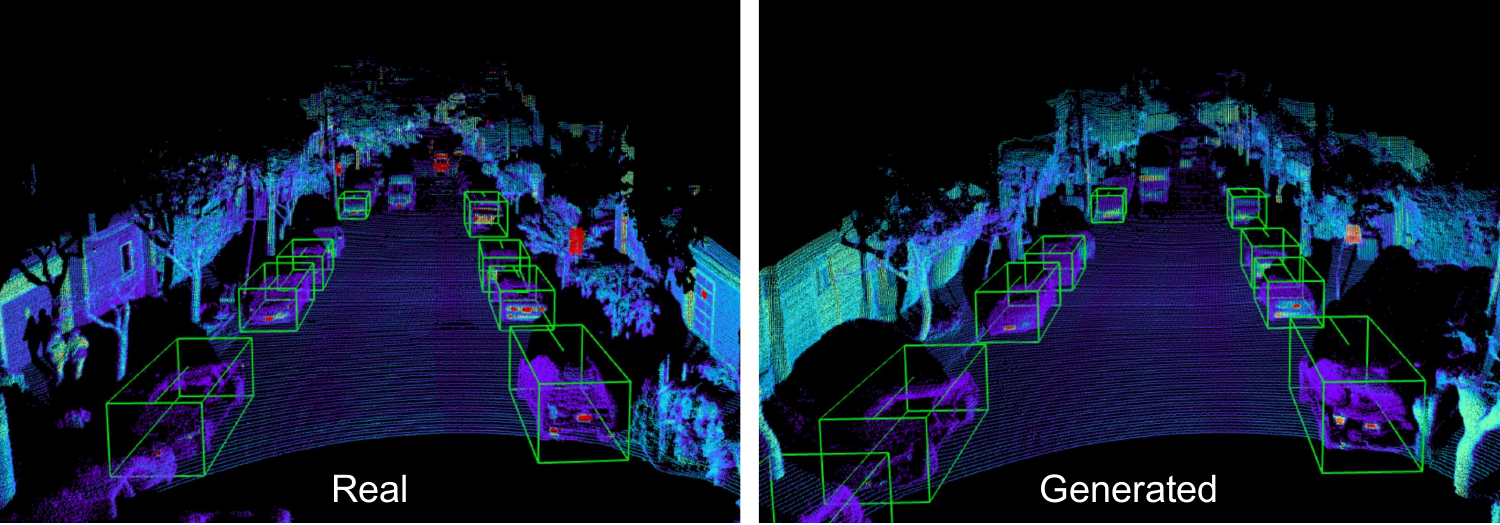}
\caption{\textbf{LiDAR detection.} We tested a vehicle detection model using real and generated LiDAR. Comparable results confirm the fidelity of our generation.}
\label{fig:lidar_detection}
\vspace{3pt}
\end{figure}

\begin{figure}[]
\centering
\includegraphics[width=\columnwidth]{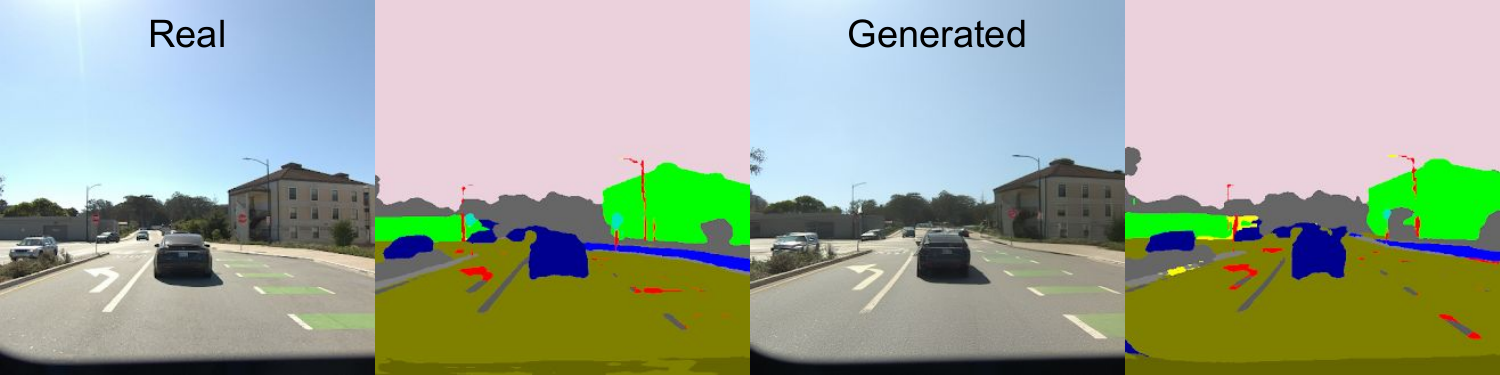}
\caption{\textbf{Image segmentation.} Panoptic-DeepLab~\cite{cheng2020panoptic} produces consistent predictions on real and generated images.}
\label{fig:image_segmentation}
\vspace{4pt}
\end{figure}

We aim to build a high-fidelity simulation environment. To assess realism, we apply perception models trained on real data directly to our generated data without finetuning. Comparable performance on real and generated data in LiDAR detection (Fig.~\ref{fig:lidar_detection}) and image segmentation (Fig.~\ref{fig:image_segmentation}) indicates strong alignment with real-world distributions.

\section{Conclusion}
\label{sec:conclusion}

\emph{\ours} is a novel generative paradigm that bridges the embodiment gap between consumer driving videos and the complex, multi-modal sensor suites required for AV validation. Leveraging a 4DGS-based data pairing pipeline and a conditional diffusion architecture, \ours converts monocular third-party videos into synchronized multi-view camera streams and LiDAR point clouds, achieving state-of-the-art performance in cross-embodiment sensor generation.
Crucially, the model co-generates consistent LiDAR and demonstrates strong generalization to real-world footage. By unlocking large-scale driving videos for AV development, our approach provides a scalable solution to data scarcity for safety-critical validation and deployment of safety-critical autonomous systems. Future work will explore improved scalability, generalization to more sensors, and a more scalable evaluation protocol.
\newpage
{
    \small
    \bibliographystyle{ieeenat_fullname}
    \bibliography{main}
}

\clearpage
\appendix
\setcounter{page}{1}
\maketitlesupplementary

\section{Extended Qualitative Results}
\begin{figure*}[t]
    \centering
    \includegraphics[width=0.72\linewidth]{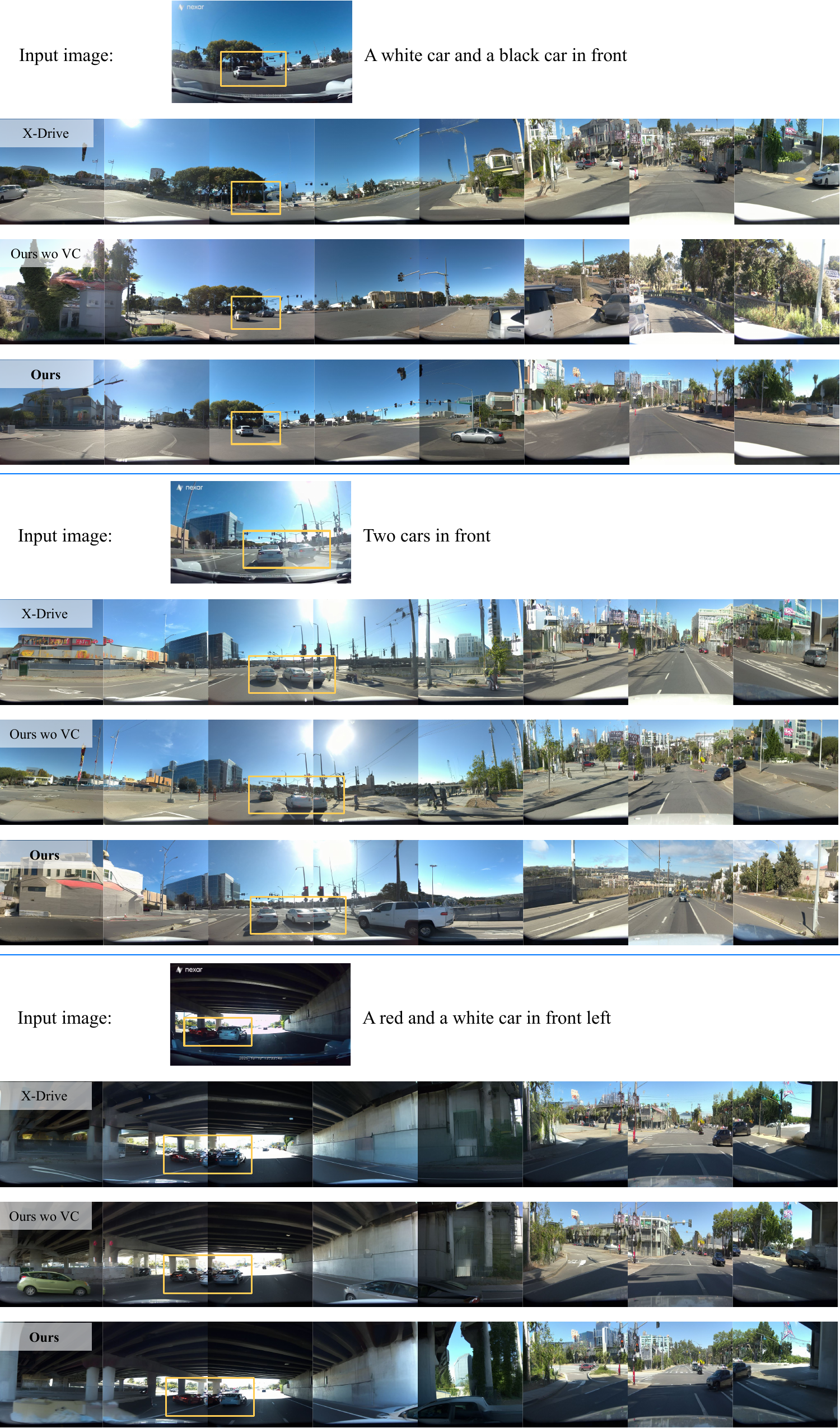}
    \caption{
        Additional qualitative results for image generation. Our proposed method demonstrates superior fidelity compared to the input dashcam image, accurately preserving the correct shape and color of objects, especially vehicles, which is challenging for the baselines.
    }
    \label{fig:image-gen1}
\end{figure*}

\begin{figure*}[t]
    \centering
    \includegraphics[width=0.72\linewidth]{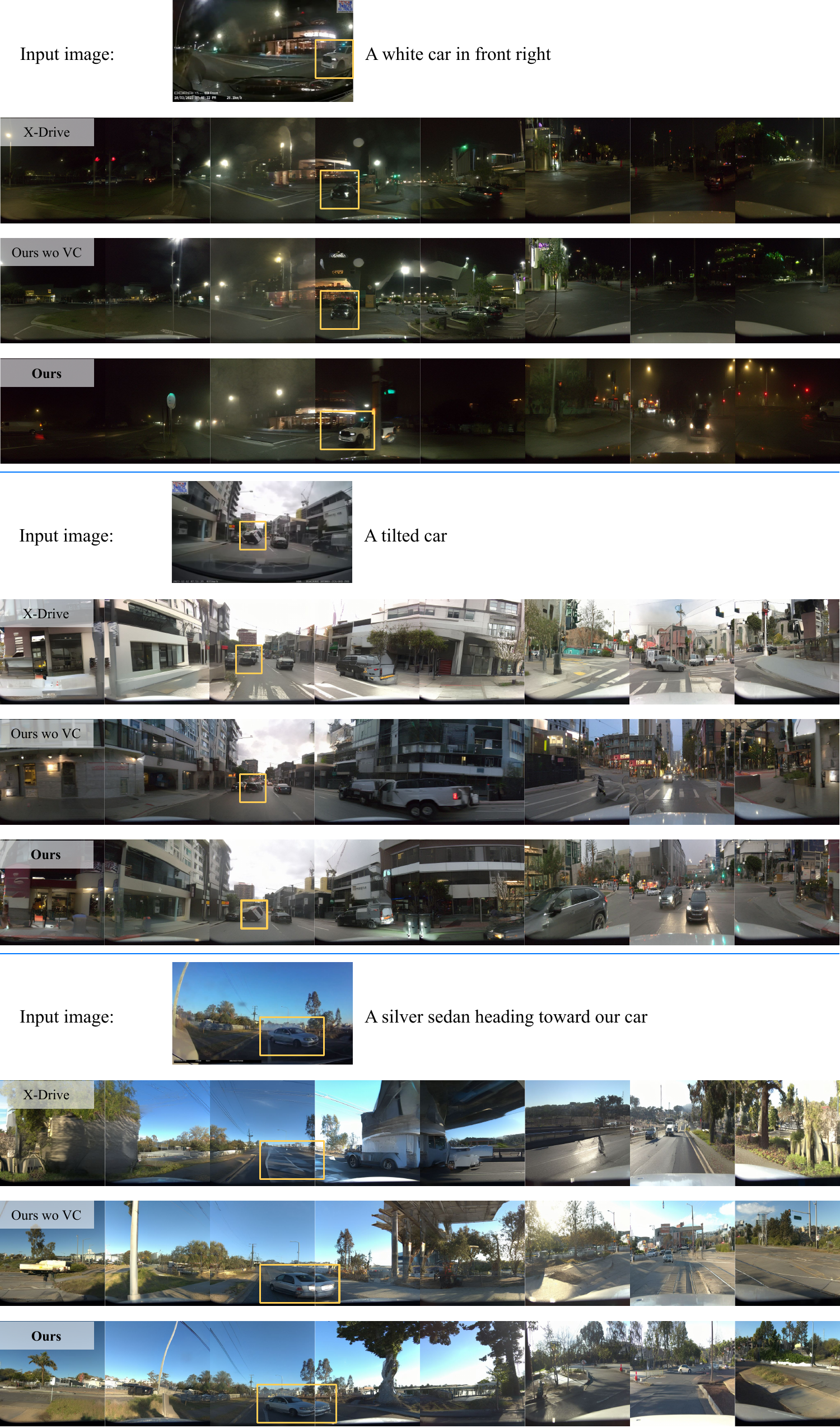}
    \caption{
        Additional qualitative results for image generation. Our proposed method demonstrates superior fidelity compared to the input dashcam image, accurately preserving the correct shape and color of objects, especially vehicles, which is challenging for the baselines.
    }
    \label{fig:image-gen2}
\end{figure*}

\begin{figure*}[t]
    \centering
    \includegraphics[width=\linewidth]{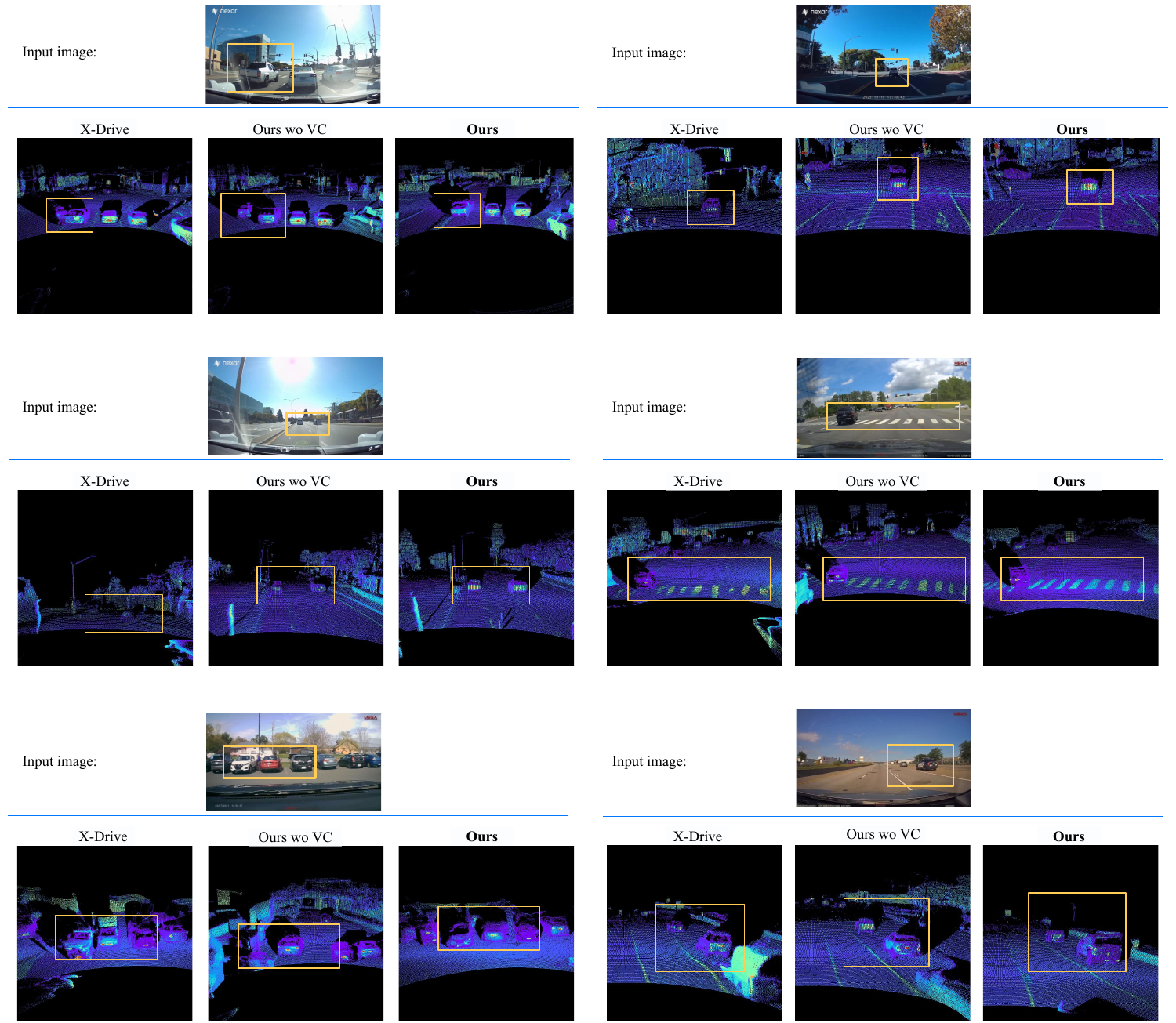}
    \caption{
        Additional qualitative results for LiDAR generation. Our method yields more accurate geometry in the synthesized point clouds, resulting in a less noisy output and a better correspondence with the accompanying image data. This improved fidelity allows for a more accurate preservation of the underlying spatial relationships of the scene.
    }
    \label{fig:lidar-gen}
\end{figure*}

\begin{figure*}[t]
    \centering
    \includegraphics[width=0.86\linewidth]{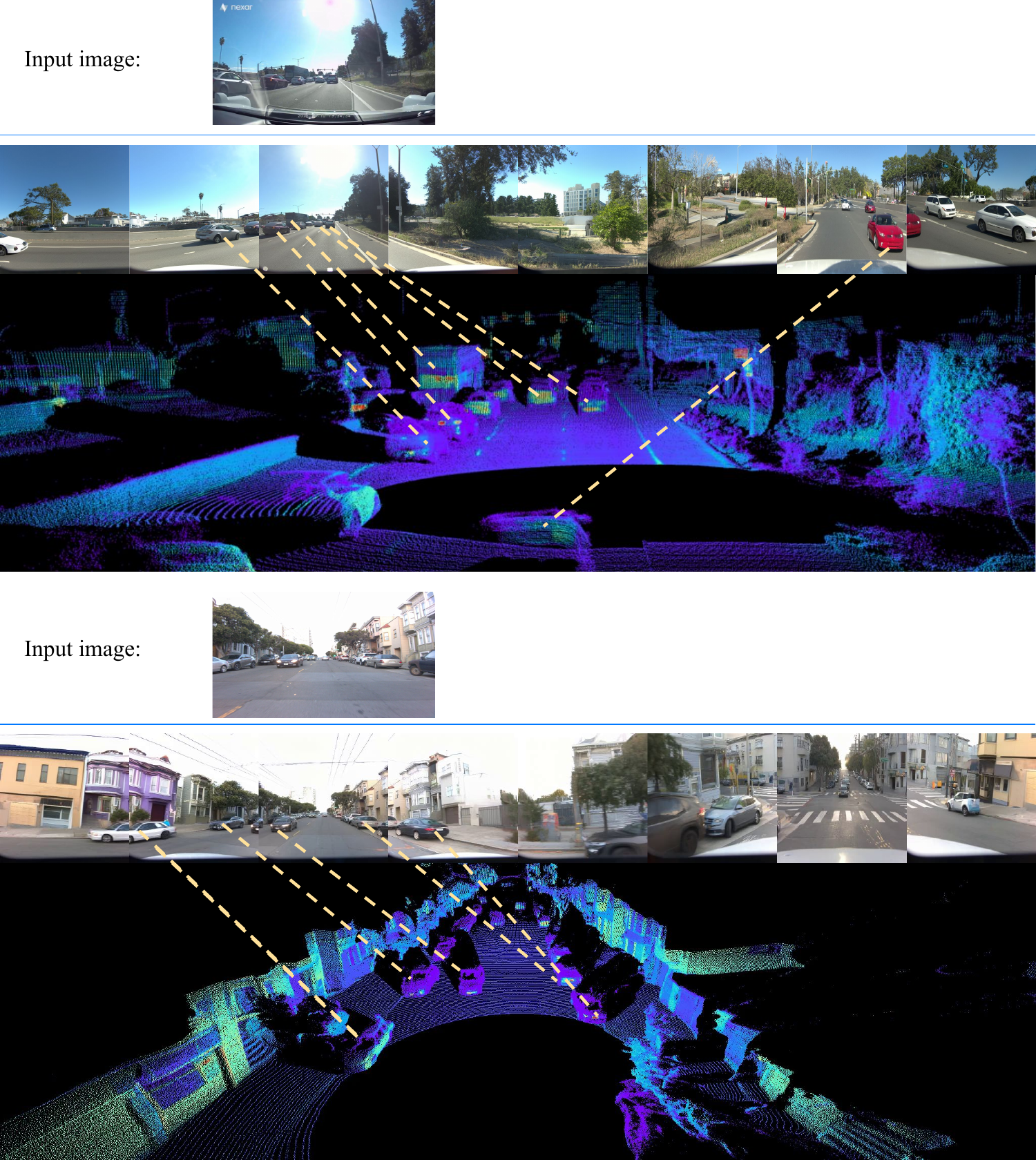}
    \caption{
        Additional qualitative results showcasing the Image-LiDAR alignment and cross-modal consistency achieved by our method. These visualizations confirm that the generated LiDAR point cloud accurately reflects the geometric details observed in the synthesized image view, demonstrating precise spatial registration between the two modalities.
    }
    \label{fig:alignment}
\end{figure*}

In this section, we provide an in-depth visual analysis to complement the quantitative results presented in the main paper. These figures are specifically designed to highlight the efficacy and generalization capabilities of our Sensor2Sensor pipeline across different output modalities.
We present additional qualitative results covering image generation (Figure~\ref{fig:image-gen1} and Figure~\ref{fig:image-gen2}), LiDAR generation (Figure~\ref{fig:lidar-gen}), and image–LiDAR alignment (Figure~\ref{fig:alignment}). 
Finally, to best illustrate the temporal coherence and realism of our full pipeline, more video generation results are presented in the accompanying \textbf{supplementary video}.

\section{Implementation Details}
\subsection{Training Pipeline} Our model is trained in a four-step pipeline to progressively incorporate increasingly complex conditioning information. \begin{itemize} \item \textbf{Step 1: Base Conditioning Single Frame Generation.} The model is first trained on single frame generation, given conditional dashcam images. \item \textbf{Step 2: Previous Frame Conditioning.} The model is then fine-tuned with dense conditioning signals, including the latent representations of the previous frame's camera and LiDAR data, as well as an additional dashcam view. \item \textbf{Step 3: DAgger Data Generation.} We use the model from Step 2 to generate a new dataset in a Dataset Aggregation (DAgger) fashion. The model is unrolled for multiple steps to create long-term simulations, which may include drifted data. \item \textbf{Step 4: DAgger Fine-tuning.} Finally, the model is fine-tuned on the DAgger-generated dataset from Step 3. This step involves training with augmented latent representations from the previous frame, which helps the model learn to correct its own errors and improves long-term simulation stability. \end{itemize}

\subsection{Model Architecture} The core of our generative model is a conditional diffusion model with a multi-stream UNet backbone designed for multi-modal sensor data.

\paragraph{Backbone.} We employ a UNet architecture with temporal attention connections. It features separate processing streams for camera and LiDAR data, allowing the model to learn modality-specific representations while fusing information through shared attention layers. The UNet processes inputs from 8 surrounding camera views and one dashcam view, along with the top-mounted LiDAR. The architecture uses a block structure with output channels of (320, 640, 1280, 1280).

\paragraph{Variational Autoencoders (VAEs) \cite{kingma2014auto}} We use separate, pre-trained VAEs to encode the raw sensor data into a compact latent space. \begin{itemize} \item \textbf{Image VAE:} A VAE~\cite{CAT3D} is used to encode the camera views into 8-channel latent representations. \item \textbf{LiDAR VAE:} A dedicated VAE encodes the raw LiDAR spin image into a 16-dimensional latent space. The UNet's LiDAR stream is configured with 16 input and output channels to match this latent space. More details about LiDAR VAE training is shown in Section~\ref{sec:lidar_vae_train}. \end{itemize}

\paragraph{Conditioning Mechanisms} The generation process is guided by conditioning inputs. \begin{itemize} \item \textbf{Dashcam Conditions:} Current frame of dashcam is conditioned into the diffusion blocks by concatenate the feature with the denoising latents in the view dimension. 
During training, we incorporate random spatial masking (with a probability of 0.2) on the conditional dashcam frames. At inference time, we can leverage this capability to apply targeted masks over distractor elements (e.g., dashcam watermarks or the ego-vehicle hood), ensuring the model focuses solely on the relevant scene context.

\item \textbf{Previous frame Conditions:} To ensure temporal consistency, the model is conditioned on the latent representations of the previous frame's camera images and LiDAR scan. We achieve this by concatenating the latents from the previous timestep to the current frame's latents along the channel dimension. Additionally, during training, we randomly drop this temporal conditioning with a probability of 0.5 to facilitate the learning of initial frame generation and improve robustness.
\end{itemize}

\subsection{Training Details} The model is trained on 128 TPUs. We use the AdamW optimizer with a learning rate of 5e-5. We clip the global norm of gradients at 1.0. For regularization, we randomly drop conditioning signals during training. For evaluation, we use an exponential moving average (EMA) of the model weights with a decay of 0.999. The training follows the multi-step pipeline described above, with each step fine-tuning from the checkpoint of the previous step. For step 1, 2, and 4, we train with 80k, 40k, and 20k steps, respectively. The number of model parameters is around 250M.

\subsection{Dataset Details} 
For training, we use a proprietary dataset of 100k 10s clips (8 cameras + top LiDAR) for 4DGS reconstruction. The resulting rendered images and synchronized sensor logs constitute the paired training data for diffusion models.
For evaluation, quantitative analysis uses 1K paired 3s sequences from proprietary Fixed-Camera-to-AV logs. For in-the-wild evaluation, we collect diverse unconstrained inputs: internet videos, ADAS logs, and manually captured dashcam (e.g., Nexar) and smartphone footage. 

\subsection{Dashcam Parameter Distribution} 
Camera parameters are sampled via a two-stage process: (1) Extrinsics: We first select a vehicle category (e.g., Sedan, SUV), then sample 6-DoF poses from category-specific distributions (e.g., for Sedans: height 1.1--1.3m, forward translation 2.0--2.5m, pitch $\pm$10$^\circ$). (2) Intrinsics: Parameters are drawn from a set of calibrated real-world dashcams (e.g., Nexar, VIOFO) and augmented with uniform noise (e.g., $\pm$5\% focal length). Final outputs undergo exposure compensation and gamma correction for lighting normalization.

\subsection{Different Target Camera Configurations} 
\textit{Sensor2Sensor} is designed for multi-sensor flexibility via its raymap-conditioning architecture, which encodes camera intrinsics and extrinsics into the generation process. While the current results focus on our large-scale proprietary platform, the raymap ensures the model is not limited to a single configuration, as it learns the fundamental mapping between 3D rays and pixel intensities. To adapt to new platforms, our paradigm simply requires 4DGS-based paired data generation for the target sensor configurations.

\subsection{LiDAR VAE Training}
\label{sec:lidar_vae_train}

We introduce a VAE architecture for generating LiDAR spin images, jointly encoding depth, intensity, and elongation. The encoder and decoder are both convolutional, and its latent space are regularized with a KL divergence loss. The normalized range, intensity, and elongation use an L1 reconstruction loss, while the validity reconstruction loss uses cross entropy. In addition, we add an LPIPS loss on surface normals (derived from predicted point cloud), intensity, elongation, and validity. The total loss, which we seek to minimize, is a weighted sum of all components, shown in Equation (\ref{eq:lidar_loss_total}), with terms:
$
\mathcal{L}^{\text{L1}}_{\text{range}} + \mathcal{L}^{\text{L1}}_{\text{elongation}} + \mathcal{L}^{\text{L1}}_{\text{intensity}} + \mathcal{L}^{\text{BCE}}_{\text{validity}} + \mathcal{L}^{\text{LPIPS}}_{\text{normals}} +
\mathcal{L}^{\text{LPIPS}}_{\text{elongation}} + \mathcal{L}^{\text{LPIPS}}_{\text{intensity}} + \mathcal{L}^{\text{LPIPS}}_{\text{validity}} + \mathcal{L}^{\text{KL}}.
$
In this formulation, the $\mathcal{L}^{\text{LPIPS}}_{\text{normals}}$ term uses normals $\mathbf{f}^L_{\text{normals}} = \texttt{ComputeNormals}(\mathbf{f}^L_{\text{range}})$ that are computed based on finite differences using the projected 3D lidar points. We now define each loss term individually.

\textbf{L1 Reconstruction Loss.}
For the signal components using L1 loss (range, elongation, and intensity), the loss is defined as:
\begin{equation}
\mathcal{L}_{\text{signal}}^{\text{L1}} = \lambda_{\text{signal}}||\mathbf{f}_{\text{signal}}^{L} - \hat{\mathbf{f}}_{\text{signal}}^{L}||_1
\end{equation}
\noindent where ``signal" represents range, elongation, or intensity. In this equation, $\mathbf{f}_{\text{signal}}^{L}$ is the ground truth LiDAR feature map and $\hat{\mathbf{f}}_{\text{signal}}^{L}$ is its corresponding reconstruction from the VAE. The term $\lambda_{\text{signal}}$ is a scalar hyperparameter that weights the contribution of this specific loss component.

\textbf{Binary Cross-Entropy Loss.}
The cross-entropy loss on the validity mask is calculated by:
\begin{equation}
\begin{split}
\mathcal{L}^{\text{BCE}}_{\text{validity}} = - \lambda_{\text{BCE}} [ &\mathbf{f}^L_{\text{valid}} \log(\hat{\mathbf{f}}^L_{\text{valid}}) \\ &+ (1 - \mathbf{f}^L_{\text{valid}}) \log(1 - \hat{\mathbf{f}}^L_{\text{valid}})]
\end{split}
\end{equation}
where $\mathbf{f}^L_{\text{valid}}$ is the ground truth binary validity mask (with values 1 for valid returns and 0 otherwise) and $\hat{\mathbf{f}}^L_{\text{valid}}$ is the predicted validity probability map output by the decoder. The $\lambda_{\text{BCE}}$ is its corresponding loss weight.

\textbf{LPIPS Perceptual Loss.}
The LPIPS (Learned Perceptual Image Patch Similarity) \cite{zhang2018perceptual} loss measures the perceptual distance between a reference image $x$ and a distorted image $\hat{x}$. Unlike traditional metrics like L1 or MSE, LPIPS leverages features extracted from a pre-trained deep neural network (e.g., VGG \cite{simonyan2014very}). The loss, presented in the equation
\begin{equation}
\mathcal{L}_{\text{LPIPS}}(x, \hat{x}) = \sum_{i} \frac{1}{H_i W_i} \sum_{h,w} \left\| w_i \odot (y^i_{hw} - \hat{y}^i_{hw}) \right\|_2^2,
\end{equation}
is computed by feeding both images through the network and calculating a weighted distance between their internal activations.
In this formulation, $i$ indexes the network layers used for the comparison. At a given layer $i$, the terms $\hat{y}^i_{hw}$ and $\hat{y}^i_{0,hw}$ represent the feature activation vectors at spatial position $(h, w)$ for images $x$ and $x_0$, respectively, which have been unit-normalized along the channel dimension. The total height and width of the feature map at this layer are given by $H_i$ and $W_i$, allowing the $\frac{1}{H_i W_i} \sum_{h,w}$ operation to compute the spatial average of the distances. The difference between activations is scaled by $w_i$, a learned channel-wise weight vector optimized to match human perceptual judgments, via the element-wise product ($\odot$). The squared L2 norm ($\left\| \cdot \right\|_2^2$) is then used to compute the distance between these weighted vectors. Finally, the total $\mathcal{L}_{\text{LPIPS}}$ is the sum of these spatially-averaged distances across all included layers $i$.

The LPIPS loss on the signals (normals, elongation, intensity, and validity) is calculated by:
\begin{equation}
\mathcal{L}^{\text{LPIPS}}_{\text{signal}} = \lambda_{\text{signal}} \mathcal{L}_{\text{LPIPS}}(\mathbf{f}^L_{\text{signal}}, \hat{\mathbf{f}}^L_{\text{signal}})
\end{equation}
Here, $\lambda_{\text{signal}}$ is the corresponding weighting factor for each specific signal type.

\textbf{KL Divergence Regularization.}
The KL divergence loss, which regularizes the latent space to follow a standard normal distribution, is calculated by:
\begin{equation}
\mathcal{L}^{\text{KL}} = \frac{1}{2} \lambda_{\text{KL}} \sum_{j=1}^{D} \left( \mu_j^2 + \sigma_{j}^2 - \log(\sigma_{j}^2) - 1 \right)
\end{equation}
This term represents the Kullback-Leibler divergence between the encoder's output distribution, $\mathcal{N}(\boldsymbol{\mu}, \boldsymbol{\sigma}^2)$, and the prior, $\mathcal{N}(\mathbf{0}, \mathbf{I})$. Here, $D$ is the dimensionality of the VAE's latent space. For each latent dimension $j$, the encoder outputs a mean $\mu_j$ and a variance $\sigma_j^2$. Finally, $\lambda_{\text{KL}}$ is the hyperparameter that balances this regularization term against the reconstruction losses.

\subsection{DAgger Training}
Dataset Aggregation (DAgger)~\cite{DAGGER} is originally an imitation learning algorithm designed to mitigate the compounding errors of behavioral cloning. It iteratively collects and aggregates data by querying an expert $\pi^*$ for optimal actions $a^*$ on states $s$ visited by the current policy $\pi_i$. A new policy $\pi_{i+1}$ is then trained on this aggregated dataset.We adapt DAgger to autoregressive video generation, treating it as a sequential decision-making process to combat temporal inconsistency.

In our case, we introduce DAgger for Video Generation. We map the components as follows.
(a) Policy $\pi$: the video generation model, which predicts the next frame.
(b) State $s_t$: the sequence of previously generated frames, $s_t = \{f_1, f_2, ..., f_t\}$.
(c) Action $a_t$: the generated next frame, $a_t = f_{t+1}$.
(d) Expert $\pi^*$: a mechanism (e.g., human evaluator, critic model, or ground-truth data) that provides a ``correct" next frame $a_t^*$ given a policy-generated state $s_t$.

First, we train a base model to generate the current frame conditioned on the ground-truth previous frame. We then utilize the base model to auto-regressively generate rollout frames for all segments in the training set. These generated frames serve as a ``degraded" dataset for augmentation. We train an improved model by randomly substituting the ground-truth history with these generated frames during training. This exposes the model to its own accumulation errors, making this model significantly more robust than the base model. While this process can be repeated iteratively, we find that a single iteration yields satisfactory rollout quality. For the DAgger training phase, we set the rollout horizon to 6 steps. Although each training segment contains approximately 35 frames, we find that training on this shorter rollout window is empirically sufficient to achieve robust performance, avoiding the computational cost of full-sequence training.

\section{Limitations and Potential Solutions}
Our approach achieves state-of-the-art per-frame generative quality, with our multi-modal diffusion model serving as a high-fidelity backbone for static scenes. We then leverage this powerful single-frame model for video synthesis by extending it auto-regressively, conditioning each new frame on the previously generated one. A limitation is, while our DAgger finetuning strategy effectively mitigates short-term error accumulation, temporal drift remains a known challenge for long-horizon sequences (e.g., $>30$ seconds). Over extended rollouts, minor prediction errors such as small geometric drifts in LiDAR or slight visual inconsistencies, can compound. This may lead to a gradual loss of long-range temporal coherence or a perceived drift in sensor calibration. However, this limitation could be addressed by incorporating a more robust longer video generative backbone designed for long-range consistency. A complementary, and more immediate, solution would be to expand the auto-regressive conditioning window. Instead of conditioning only on the single prior frame ($t-1$), the model could attend to a richer temporal context (e.g., $t-k, ..., t-1$). This would provide stronger priors for maintaining object-level and scene-level consistency over time. While outside the primary scope of this work, we leave these promising directions for future exploration.

\section{Synthetic Cameras from 4DGS}
\begin{figure}
    \centering
    \includegraphics[width=\linewidth]{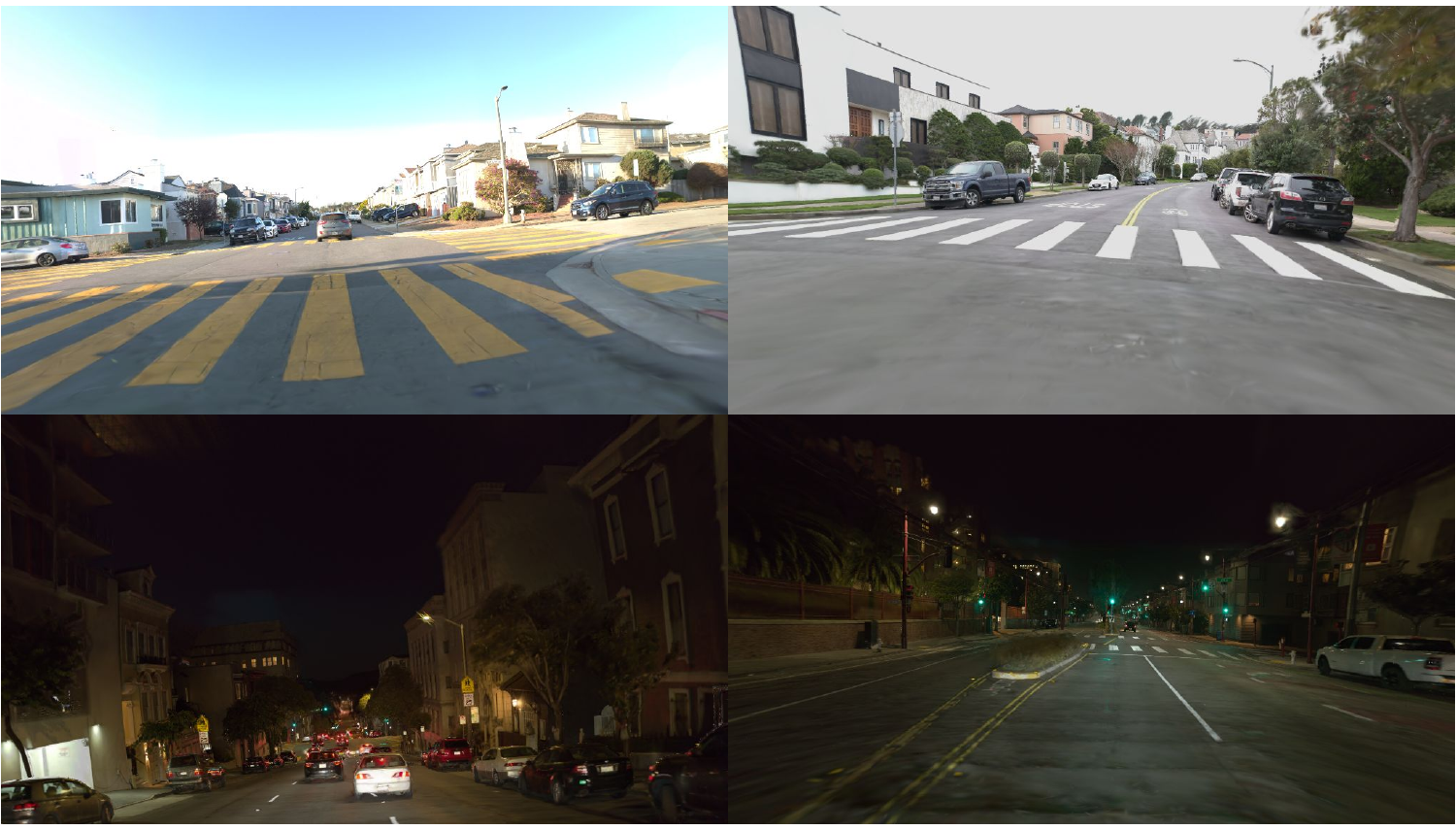}
    \caption{
        Visualization of synthetic dashcam images rendered from 4DGS across diverse camera settings. The renders demonstrate high visual fidelity and realism, effectively simulating the characteristics of in-the-wild footage used for training.
    }
    \label{fig:supp-4dgs}
\end{figure}

One important component of our pipeline is the utilization of 4D Gaussian Splatting (4DGS) to synthesize paired training data by simulating third-party camera views. As illustrated in Figure~\ref{fig:supp-4dgs}, the synthetic dashcam images rendered via our 4DGS pipeline exhibit high photorealism, faithfully mimicking the optical characteristics and environmental complexity of real-world dashcam footage.

Crucially, our diffusion model is trained to map these synthetic inputs ($I_{synth}$), which may contain minor reconstruction artifacts such as floaters or slight blur, to pristine, ground-truth real sensor data ($O_{real}$). This training objective effectively functions as a denoising task, forcing the network to learn robust spatial and semantic mappings between the monocular view and the target sensor suite, rather than overfitting to low-level input artifacts. Consequently, at inference time, the model demonstrates significant robustness when presented with sub-optimal or noisy real-world dashcam inputs, successfully generating coherent and geometrically consistent AV logs.

\end{document}